# Robot to support older people to live independently

Sara Cooper
PAL Robotics
Barcelona, Spain
sara.cooper@pal-robotics.com

Óscar Villacañas
Clínica Humana
Mallorca, Spain
oscar.villacanas@pal-robotics.com

Luca Marchionni
PAL Robotics
Barcelona, Spain
luca.marchionni@pal-robotics.com

Francesco Ferro
PAL Robotics
Barcelona, Spain
francesco.ferro@pal-robotics.com

**ABSTRACT**

This paper presents an overview on how the PAL Robotics ARI robot is participating in the European SHAPES project to promote healthy and active living among older people, by integrating digital solutions from project partners and adapting the system in order to improve human-robot interaction and user acceptability in a wide range of tasks.

## 1  Introduction: SHAPES project

Older adults prefer to promote healthy ageing at home, and so it is essential to focus efforts on developing new assistive technologies [1]. In the past few years, social robots have been gaining considerable importance, with projects and social robots focusing on elderly care (Companionable [2], Care-o-Bot [3], Hobbit [4], Pepper [8]). However, deployment in real-world scenarios is still uncommon and, consequently, the level of acceptance of such solutions remains controversial.

The SHAPES [1]project [5] aims to create the first European open Ecosystem to enable the large-scale deployment of a broad range of digital solutions for supporting and extending healthy and independent living among older individuals. This is becoming even more necessary now due to the ongoing COVID-19 pandemic [6], in which social isolation and loneliness greatly impact older people.

One of the means available is through the use of social companion robots, which can assist older people in a humanised social way to reinforce engagement. As part of Pilot 1 - Smart Living Environment for healthy ageing, out of 7 Pilots, this paper presents an overview on how PAL Robotics' ARI robot is being adapted [7].

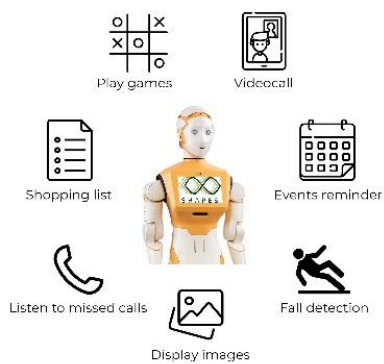

**Figure 1: PAL Robotics ARI robot to be equipped with multiple functions to promote healthy living**

ARI [2] is a high-performance robotic platform designed for a wide range of multimodal expressive gestures and behaviour, focusing on social interaction. ARI is provided with a mobile base, a torso with an integrated Linux-based tablet, two arms and a head with expressive gaze. It provides natural and intuitive multimodal interaction via its touchscreen, speech, light-emitting diodes and gestures. With a 4-microphone board on the torso to detect speech, a set of RGB-D and RGB cameras for environment detection, and autonomous behaviour, it is capable of perceptive and speech interaction abilities to enable it to interact with its surroundings. The added value of ARI's prototype through SHAPES makes it an easy addition to partner modules that will add advanced chatbots to the robot, speech recognition, emotion recognition, fall detection, and videocalling capabilities, to name a few examples.

As a social robot, ARI can be used as a therapeutic assistant in hospitals, carehomes or end-user homes to foster social communication, reduce loneliness, stress, and increase overall user enjoyment and activity.

The participation of ARI in pilot 1 will take place at Clínica Humana [3] (Mallorca, Spain). This is a private clinic with more that 7-years' expertise in chronic patient management (currently, over 600 patients), and they provide hospital care to retirement homes, communities and homebound patients that have a major technological component in the form of telemedicine. The pilot test consists of different stages, with the first one, the mock-up session, involving a presentation of the pilot test as well as the robot's capabilities to all actors in a real scenario. The purpose of the mock-up is to better understand which of the proposed activities, as shown in Figure 1, are of most interest to end-users, the degree of interaction they would like to have with the robot, and to evaluate their initial perception. At a later stage, hands-on training will follow prior to actual pilot testing.

---

[1] https://shapes2020.eu/
[2] https://pal-robotics.com/robots/ari/
[3] https://www.clinicahumana.es/

**2 User requirements gathering**

End-users are among people between 70-80 years old, living in urban environments. These users live in their houses or in sheltered housing and may receive regular visits, mainly from their family members. The pilot test will be conducted with 4-5 participants for 4 weeks each. In order to develop the prototype, a human centred co-design process has been applied, by first defining the needs of the participants through the help of Clínica Humana: mainly older people, caregivers and general practitioners, but also desk staff at the sheltered apartment complex and other technicians. Tasks to be implemented have been selected from multiple interactions with these users, as well as considering the platform user experience design guidelines developed as part of the SHAPES project to allow for positive user experience, validated by the SHAPES ecosystem regarding relevant stakeholders.

In order to remain independent, older people often need assistance to remember appointments, take medication, keep physically active and also be further engaged socially in order to reduce risk of isolation. Moreover, the COVID-19 pandemic has also resulted in additional needs such as temperature monitoring.
Based on user requirements, the robot will be programmed to carry out some of the following actions, summarised in Figure 1:
- Detect and monitor temperature
- Establish a video call
- Send alerts to caregivers by call, SMS or other means about incidents
- Receive messages from caregivers and third parties
- Detect and monitor falls
- Remind users of different events
- Prompt follow-ups
- Provide entertainment games and physical exercise games
- Show pre-selected images on the touchscreen
- Set up shopping lists and remind users to fill in the shopping list

During the mock-up sessions, the type of reminders they would like to have, how often, and the types of alerts caregivers would like to receive, will be gathered.

In addition, the system will provide the option for the health-care personnel and/or the caregiver to receive alerts from the robot that enable monitoring the end-user as well as to establish videocalls, provided by the SHAPES platform.

**3 Prototype development and multi-modal interaction**

The main consideration in prototype development is the design of a multimodal user interface, a necessity for the robot's successful deployment in human environments and effective interaction [8]. Such social cues will also directly affect how end-users interpret and understand robot intentions [9].

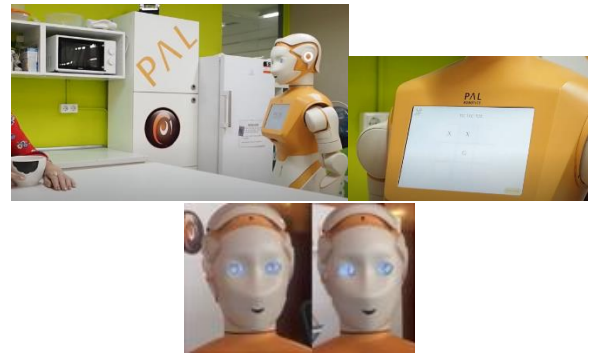

**Figure 2: Types of human-robot interaction**

The robot offers visual, voice and tactile interaction modes via the following means, some of which are shown in Figure 2:

- Touch-screen interaction (Figure 2): display static images, videos, slideshows (series of images), buttons that trigger other robot actions (with adjustable sizes and colours for increased usability), and HTML packaged content (e.g. Javascript games)

- Speech interaction: text-to-speech in multiple languages, with possibility to adjust speed; as well as Google Cloud API for speech recognition. There is the option to group a variety of sentences per topic (e.g. presentation, goodbyes) and output a randomly selected option. This is useful in this user-case, for example when ARI wants to engage in an activity with the user, to produce a set of different phrases like "Hey, do you want to play with me?"; "Hello, how are you today?; Would you like to do something?".

- LED effects in both ears and back torso: adjustable colours, option to synchronise all LEDs or adjust individually. Some effects include blinking, and fading from one colour to another.

- Gestures: the robot has 4 degrees of freedom (DoF) arms and 2 DoF head, allowing it to adjust visual proxemic and kinesic perception. New motions can be created using the motion builder [4], shown in Figure 3, enabling non-technical users to create or adjust already created motions by using the robot's Web GUI. This is useful if during hands-on training they consider the need to reduce the speed of some motions to ensure greater perceived safety.

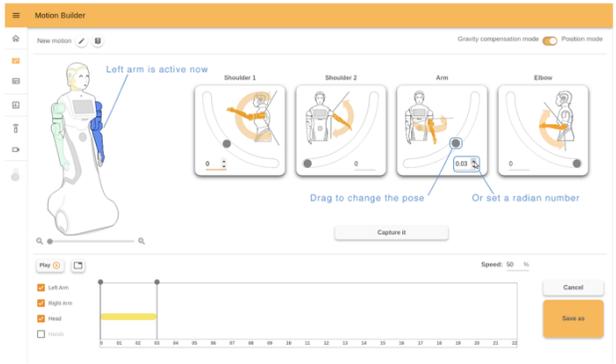

**Figure 3: Web GUI motion builder to generate new motions**

ARI has some default behavioural cues such as waving, shaking hands and nodding. Text-to-speech is combined with created gestures to achieve animated speech.

- Expressive animated eyes (Figure 2): whenever the robot is switched on, it will have autonomous gaze behaviour where it moves the pupils randomly. At present, this behaviour cannot be adjusted, although it will be a matter of interest for the future.

In order to synchronize these different types of interaction, PAL Robotics provides the Speech Editor, with a screenshot of a version of it shown in Figure 4, which is a tool to create a new robot presentation combining speech, LED and touch-screen content. Using ROS, integrators can also synchronize these components in a single command.

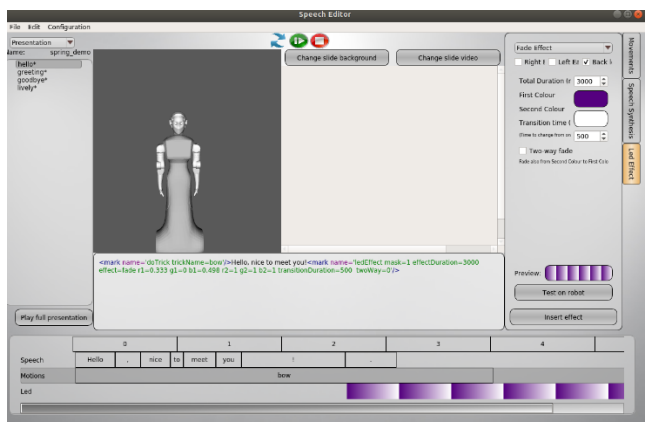

**Figure 4: PAL Robotics Speech Editor**

---

4        http://wiki.ros.org/play_motion_builder

As an example, in one of the entertainment games, the robot will provide instructions for the game using animated speech, and also provide further encouragement as the user plays using the touchscreen. In any case, once the robot is deployed at the pilot site between May-June 2021, and hands-on training begins, a more in-depth evaluation on which specific type of social cues will be used for each task will be evaluated, as well as assessment on which ones are more user-preferred.

While touch-screen interaction provides greater reliability, speech and gesture-based interaction are more natural means of communication for most people [4]. Furthermore, speech interaction is considered to be a challenge, especially among older people [10], not to mention the fact that many such older people lack sufficient familiarity with new technologies to comfortably handle a touch-screen or tablet or may have visual impairments. Usability, accessibility and acceptability parameters will be analysed thanks to hands-on training in the initial stages of the pilot test, which will allow time for adjustments. To increase accessibility, a second tablet will be integrated in the back part of ARI, and this may be retrieved by end-users so that it is easier to interact with if they are seated, as shown in Figure 5. Videocalling will be evaluated, using both touchscreens to assess user preference. Apartments of different users are also expected to be different, with some being tidier than others - this will also lead to challenges during navigation.

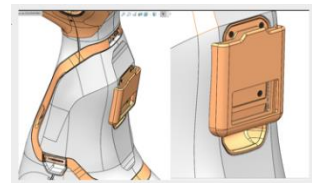

**Figure 5: Integration of additional back tablet**

Additional modifications to the robot include the inclusion of a thermal camera, especially in response to the COVID-19 pandemic, so that the robot may measure the older person's temperature and alert caregivers if the temperature is abnormal. Digital solutions from several partners of SHAPES will be integrated into the robot. To do so, ARI offers its Rest API so that digital solutions may call its functions externally, enabling use of other programming languages (e.g. Java, Android). Moreover, digital solutions may be integrated into the robot as ROS (Robotics Operating System) nodes.

These SHAPES solutions will enhance the robot's interaction skills. For instance, the robot will be able to detect user engagement and emotions, and decide whether to continue or stop the interaction, switch from one game to another, or initiate an interaction (e.g. if the user appears bored). It will also be integrated with a new Automatic Speech Recognition engine and a chatbot, to enhance its speech interaction. Additionally, there is a fall detection mechanism, videocalls and user authentication using multimodal biometrics, to enable user data to be accessed from the SHAPES Platform [11].

There will be two main methods of interaction and for selecting the task to be performed: in the first case, the older person will approach the robot and ask to begin the interaction using the touchscreen or speech (e.g. "Hello, ARI, can you tell me when my appointment is?'"; "Hello, ARI, can we play the game Tic Tac Toe?"). In the second case, the robot will proactively search for the user in the apartment according to the reminders set by the caregivers (it could navigate from the living to the kitchen, where the user is, as shown in Figure 2), once a specific time frame has elapsed since the last interaction and it detects a user close by. In all cases, ARI will promote the interaction by using its multi-modal behavioural skills mentioned previously.

The prototype will be developed by technical partners from the consortium, and end-users are not expected to contribute to its programming. However, ARI's Web GUI, web application to control the robot will be provided to monitor the robot's status, such as battery level (Figure 6), monitor the older person if necessary via the robot's cameras - in the case of an emergency-, or upload new images to the robot touchscreen - so that caregivers may introduce the preferred set for each user.

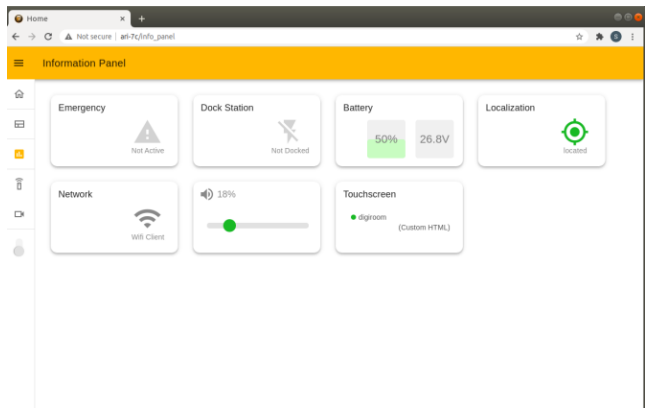

**Figure 6: Web GUI for robot status monitoring**

Based on the feedback from the hands-on training, some aspects that will be modified include, but are not limited to:
- Amount of encouragement of feedback the robot gives during activities
- Locations where the robot travels to within the house
- Human-robot proxemics to establish a user-preferred interaction distance or robot position relative to the user

## 4 Conclusions

This paper presents the SHAPES project and an overview on how the PAL Robotics ARI robot will be used to assist older people in their homes.
A user-centered design approach has been taken to gather user requirements via the use of mock-ups so as to define tasks such as offering reminders, videocalls and games. The robot's key multi-purpose aspects have been outlined, such as speech, gestures and touchscreen, as well as prototype design considerations. Subsequent hands-on training and pilot activities with the different end-users involved will further enable us to adapt human-robot interaction and robot behaviour.

**ACKNOWLEDGMENTS**

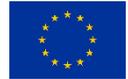

This work was supported by the SHAPES project, which has received funding from the European Union's Horizon 2020 research and innovation programme under grant agreement no. 857159